\documentclass[runningheads]{llncs}
\usepackage[T1]{fontenc}
\usepackage{graphicx}
\usepackage{amsmath}
\usepackage{subfig}
\usepackage{amssymb}
\usepackage{float}
\usepackage{booktabs}
\usepackage{multirow}
\usepackage{multicol}
\begin{document}
\title{Decoding the Stressed Brain with Geometric Machine Learning}
%
%


\author{Sonia Koszut\inst{1} \and
Sam Nallaperuma-Herzberg\inst{1} \and
Pietro Lio\inst{1}}

\authorrunning{S. Koszut, S. Nallaperuma-Herzberg, and P. Lio}

\institute{
  Department of Computer Science and Technology, University of Cambridge, Cambridge, UK\\
  \email{smk79@cam.ac.uk}
}

\maketitle              
\begin{abstract}
Stress significantly contributes to both mental and physical disorders, yet traditional self-reported questionnaires are inherently subjective. In this study, we introduce a novel framework that employs geometric machine learning to detect stress from raw EEG recordings. Our approach constructs graphs by integrating structural connectivity (derived from electrode spatial arrangement) with functional connectivity from pairwise signal correlations. A spatio-temporal graph convolutional network (ST-GCN) processes these graphs to capture spatial and temporal dynamics. Experiments on the SAM-40 dataset show that the ST-GCN outperforms standard machine learning models on all key classification metrics and enhances interpretability, explored through ablation analyses of key channels and brain regions. These results pave the way for more objective and accurate stress detection methods.

\end{abstract}

\section{Introduction} 
Stress, defined as a state of imbalance when demands exceed an individual’s adaptive capacity \cite{mcewen1998stress}, is increasingly recognised as a major causal factor in both mental and physical disorders. Traditional diagnostic methods rely on self-reported questionnaires such as the Perceived Stress Scale (PSS) \cite{cohen1994perceived}, which may not capture the details required for an accurate and fully objective assessment. Advances in non-invasive neuroimaging, particularly Electroencephalography (EEG), offer a promising means to accurately detect rapid brain function changes associated with stress. Conventional EEG processing pipelines include band-pass filtering, artefact removal, and manual spectral-feature engineering (e.g., power spectral density calculation), followed by a classifier. 

Recent deep-learning methods can differentiate stressed from non-stressed EEG \cite{malviya2022novel,patel_stress_eeg_rnn,MANE2023100231_lstm_cnn_stress_eeg}. Models such as Convolutional Neural Networks (CNNs) and Long Short-Term Memory (LSTM) networks have demonstrated the potential to extract temporal features from EEG signals; however, they typically treat the data as flat time series or grid-like arrays, ignoring the anatomical and functional connectivity among electrodes. Recent advances in geometric machine learning applications in modelling the EEG signal for emotion recognition \cite{li2022semi,gao_eeg_gcn_emotion,GUO2024128445_EEG_GNN_emotion} have sparked motivation to apply similar architectures for stress diagnosis. Graph-based methods offer a natural framework for incorporating spatial relationships in EEG.

In this study, we propose a novel framework that leverages geometric machine learning (GML) to model stress-related brain dynamics by transforming temporal EEG signals into graph representations. Although spatio-temporal GML has been applied to EEG data in other cognitive and neurological contexts \cite{shan2022spatial,li2024temporal}, to our knowledge, it has not yet been explored for stress detection. 

Our method constructs graphs from EEG recordings by combining structural and functional connectivity between electrodes, which are then processed by a spatio-temporal graph-convolutional network (ST-GCN) to capture both spatial correlations and the temporal evolution of the signal. To build each adjacency, we construct a structural graph based on inter-electrode distance \cite{bullmore2009complex_brain_graph} and a functional graph derived from correlation thresholds \cite{rubinov2010complex_brain_connectivity_struct_funct}. We then fuse these graphs via element-wise averaging, yielding a single hybrid adjacency. Our approach improves classification metrics over the best performing, Transformer-based benchmark model, by 8 percentage points in accuracy and 7 percentage points in F1 score. Moreover, it provides improved interpretability, explored through ablation studies that reveal the relative contributions of specific channels, brain regions, and temporal segments.

Our study advances EEG-based stress detection and offers new insights into relevant EEG biomarkers. The remainder of this paper is organised as follows. Section~\ref{sec:background} reviews the current state of EEG analysis and machine learning techniques. Section~\ref{sec:approach} introduces the graph construction and model architecture. Section~\ref{sec:set-up} describes the experimental setup, and Section~\ref{sec:results} presents the results. Finally, Section~\ref{sec:conclusion} summarises our findings and outlines future research directions.

\section{Background}
\label{sec:background}
\subsection{Electroencephalography}

\begin{figure}
    \centering
    \includegraphics[width=0.5\linewidth]{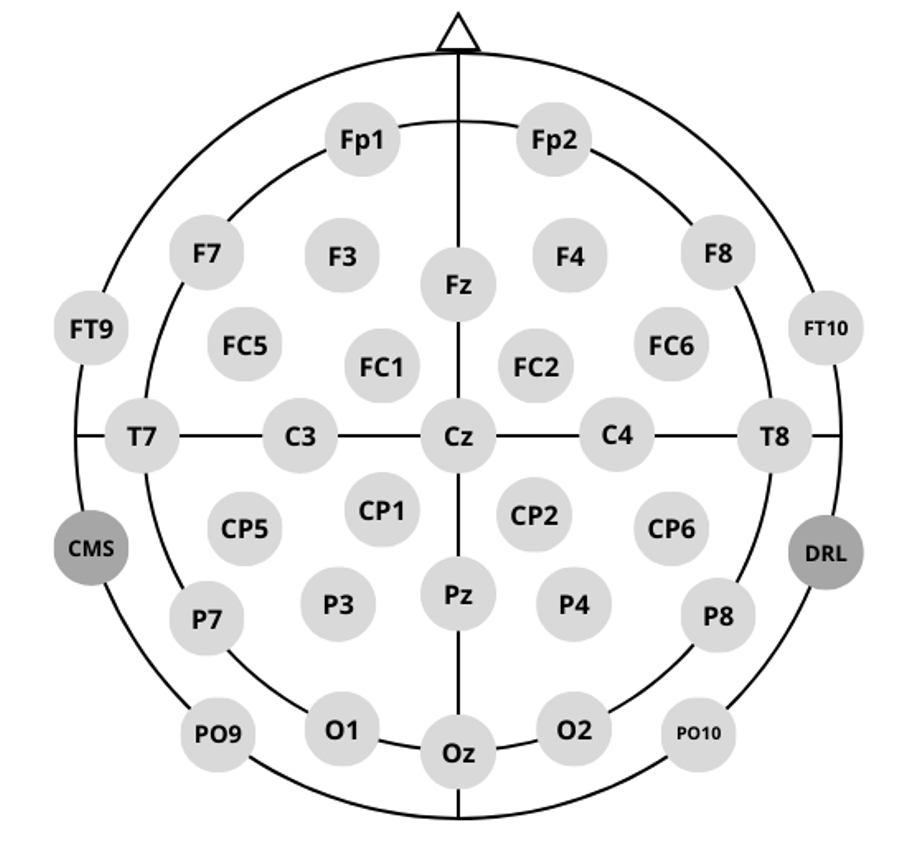}
    \caption{EEG electrode placement according to the 10-20 system \cite{jasper1958ten_10_20_system}.}
    \label{fig:eeg-10-20}
\end{figure}

Electroencephalography (EEG) is a non-invasive technique that measures the brain's electrical activity via electrodes placed on the scalp. These electrodes detect voltage fluctuations resulting from ionic currents flowing in the brain, capturing neural oscillations across various frequency bands. The International 10–20 system \cite{jasper1958ten_10_20_system} standardises electrode placement by positioning electrodes at intervals of 10\% or 20\% of specific cranial measurements, ensuring consistency and reproducibility in EEG recordings (see Figure \ref{fig:eeg-10-20}).

EEG's high temporal resolution enables the real-time monitoring of neural dynamics, making it particularly suitable for detecting rapid changes associated with stress responses. It is cost-effective and portable, which contributes to its widespread use in both clinical and research settings \cite{amol_s_chaudhari_2024_eeg_non_invasive}.

\subsection{Machine Learning for EEG-Based Stress Detection}

In recent years, machine learning (ML) has significantly advanced EEG-based stress detection by extracting complex patterns from neural signals. Various ML models have been proposed, each with different approaches to signal preprocessing and feature extraction. A comparative study \cite{sundaresan2021evaluating} tested a Convolutional Neural Network (CNN) approach and a Long Short-Term Memory (LSTM) approach, with the two-layer LSTM achieving the best accuracy. Further studies \cite{kaminska2021detection} found a CNN model to be outperformed by simpler architectures like a Multilayer Perceptron (MLP) and a Support Vector Machine (SVM).

LSTMs are capable of capturing temporal dependencies in sequential data, which makes them suitable for EEG analysis. LSTMs and hybrid LSTM-based models have been very effective in stress diagnosis \cite{malviya2022novel}. A Bidirectional LSTM applied to EEG processing for mental workload estimation has shown an over 80\% accuracy on some datasets and improved the efficiency of Brain-Computer Interface (BCI) systems \cite{chakladar2020eeg}. Transformer models have recently been explored in EEG processing for the improved capability to handle long-range dependencies. Their versatility and strong performance have been proven in various settings, including for stress modelling \cite{vafaei2025transformers,siddhad2025neural}. 

However, while these models are effective in processing temporal patterns, they do not leverage the spatial relationships in EEG. This limitation is relevant for stress detection, where certain brain regions are known to be associated with stress responses \cite{hag2023mental_eeg_biomarkers,alshorman2022frontal}. Spatio-temporal graph neural networks (ST-GNNs) have recently been applied to emotion recognition \cite{liu2024graph_emotion} and seizure detection \cite{tang2021self}, demonstrating improved classification performance and interpretability.

This study introduces a novel graph-based framework for classifying stress from raw EEG data using a ST-GCN architecture. Beyond classification performance, we are interested in the interpretability of the model. By analysing the contribution of individual channels, brain regions, and time segments through ablation studies, we aim to identify neural biomarkers of stress and EEG electrodes that are particularly important for recording stress responses. This study bridges the gap in existing stress detection methods and contributes to the neuroscientific understanding of stress.

To our knowledge, this is the first work to investigate a ST-GCN for stress detection on raw EEG. We also provide interpretability by analysing the contribution of individual channels, brain regions and temporal segments through ablation. This bridges a gap in existing methods and advances our neuroscientific understanding of stress.

\section{Approach}
\label{sec:approach}

\subsection{Graph Construction}

To capture the spatial structure of EEG recordings, each trial is transformed into a graph $G = (V, E)$, where the set $V$ represents the electrodes and $E$ contains edges defined by a hybrid measure that combines structural and functional connectivity.
 
Electrode positions are provided in a standard coordinate space. The structural connectivity is defined by the inverse of the Euclidean distance between electrode positions. Formally, let the electrode positions be denoted by $\mathbf{p}_i \in \mathbb{R}^2$ for $i=1,\dots,N$ (with $N=32$). Then, the structural connectivity matrix $\mathbf{A}_{\text{struct}}$ is given by:
\begin{equation}
\mathbf{A}_{\text{struct}}(i,j) = 
\begin{cases}
\frac{1}{\|\mathbf{p}_i - \mathbf{p}_j\|_2 + \epsilon}, & \text{if } j \in \mathcal{N}_k(i), \\
0, & \text{otherwise,}
\end{cases}
\label{eq:structural}
\end{equation}
where $\mathcal{N}_k(i)$ denotes the set of $k$ nearest neighbours of node $i$ (with $k=2$ in our experiments) and $\epsilon$ is a small constant to avoid division by zero \cite{bullmore2009complex_brain_graph}.

Functional connectivity is computed using the Pearson correlation coefficient between the raw EEG signals of different channels. Let $\mathbf{x}_i \in \mathbb{R}^{3200}$ be the normalised EEG time-series from electrode $i$. The functional connectivity matrix $\mathbf{A}_{\text{func}}$ is then defined as:
\begin{equation}
\mathbf{A}_{\text{func}}(i,j) = \mathbb{I}\Big(\rho(\mathbf{x}_i,\mathbf{x}_j) > \tau\Big),
\label{eq:functional}
\end{equation}
where $\rho(\mathbf{x}_i,\mathbf{x}_j)$ is the Pearson correlation coefficient between $\mathbf{x}_i$ and $\mathbf{x}_j$, $\tau=0.5$ is the correlation threshold, and $\mathbb{I}(\cdot)$ is the indicator function \cite{rubinov2010complex_brain_connectivity_struct_funct}.

The final connectivity matrix $\mathbf{A}$ used to define the graph is computed by combining $\mathbf{A}_{\text{struct}}$ and $\mathbf{A}_{\text{func}}$. In our approach, we perform an element-wise average:
\begin{equation}
\mathbf{A} = \frac{\mathbf{A}_{\text{struct}} + \mathbf{A}_{\text{func}}}{2}.
\label{eq:combine}
\end{equation}
This design choice combines both the anatomical proximity and the dynamic functional relationships between electrodes.

To justify the choice of graph construction parameters, we computed graph quality metrics such as the average degree, clustering coefficient, and algebraic connectivity for graphs constructed with $k=2$ nearest neighbours and connectivity threshold=0.5. The averaged results are presented in Table \ref{tab:graph_metrics}, demonstrating that the resulting graphs are well-connected, achieve a high algebraic connectivity, moderate clustering, strong structural coherence and functional grouping. 

\begin{table}[]
    \centering
    \caption{The average graph quality metrics of a constructed connectivity graph with $k=2$ nearest neighbours and connectivity threshold=0.5.}
    \begin{tabular}{|c|c|c|c|}
        \hline
         Algebraic connectivity & Clustering & Avg shortest path & Avg degree \\
         \hline
         0.85 & 0.39 & 0.55 & 20.0 \\ 
         \hline
    \end{tabular}
    \label{tab:graph_metrics}
\end{table}

To evaluate the impact of these parameters on classification performance, we performed a sensitivity analysis by varying $k \in {2, 3, 4}$ and threshold $\tau \in {0.4, 0.5, 0.6}$ in graph construction and recorded the accuracy of the ST-GCN model with such settings and a fixed random seed. The results are demonstrated in Table \ref{tab:param_sensitivity}. The resulting changes in model accuracy were not large (less than 6\% variation), suggesting that the ST-GCN is robust to moderate changes in the graph construction process. The choice of $k = 2$ and $\tau = 0.5$ was selected as it yielded the best average test performance without significantly increasing graph density.

\begin{table}[]
    \centering
    \caption{Model accuracy (rounded to the nearest integer) for different combinations of $k$ nearest neighbours and connectivity threshold $\tau$.}
    \label{tab:param_sensitivity}
    \begin{tabular}{|c|c|c|c|c|c|c|c|c|c|}
        \hline
        $k$ & \multicolumn{3}{|c|}{2} & \multicolumn{3}{c|}{3} & \multicolumn{3}{c|}{4} \\
        \hline
        $\tau$ & 0.4 & 0.5 & 0.6 & 0.4 & 0.5 & 0.6 & 0.4 & 0.5 & 0.6 \\
        \hline
        Accuracy (\%) & 72 & 78 & 73 & 74 & 75 & 72 & 74 & 72 & 73 \\
        \hline
    \end{tabular}
\end{table}

\begin{figure}[]
    \centering
    \includegraphics[width=0.5\linewidth]{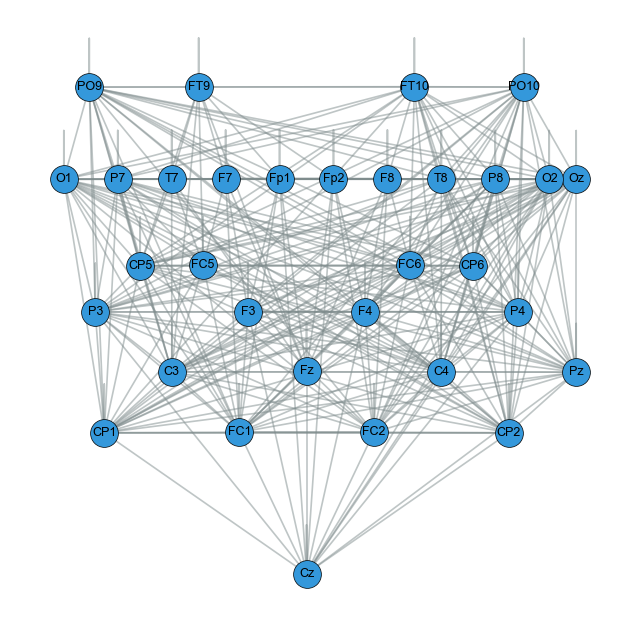}
    \caption{Example connectivity graph computed from averaged structural and functional metrics using $k=2$ nearest neighbours and a correlation threshold of $0.5$.}
    \label{fig:graph}
\end{figure}

A schematic representation of an example connectivity graph is shown in Figure~\ref{fig:graph}.

\subsection{Spatio-Temporal Graph Convolutional Network (ST-GCN)}

The ST-GCN model is designed to classify EEG signals by simultaneously learning from their geometric structure and temporal dynamics. Each trial is represented as a 3D tensor $\mathbf{X} \in \mathbb{R}^{N \times T \times 1}$, where $N=32$ is the number of channels and $T=3200$ is the number of time steps.

The model architecture consists of two main stages: temporal feature extraction and spatial message passing.

\subsubsection{Temporal Feature Extraction:}  
A single \textit{TimeDistributed} 1D convolutional layer is applied independently to the time-series data of each electrode-node, learning local temporal patterns. Let $\mathbf{X}_i \in \mathbb{R}^{T \times 1}$ denote the signal at electrode $i$. The convolution operation is defined as:
\begin{equation}
\mathbf{h}_i = \sigma\big( \mathbf{W}_c * \mathbf{X}_i + \mathbf{b}_c \big),
\end{equation}
where $\mathbf{W}_c$ and $\mathbf{b}_c$ are the learned kernel and bias, and $\sigma(\cdot)$ denotes a non-linear ReLU activation. Global average pooling is then applied along the time dimension to produce per-node feature embeddings of dimension $F$:
\begin{equation}
\mathbf{z}_i = \frac{1}{T} \sum_{t=1}^{T} \mathbf{h}_i[t].
\end{equation}
This produces a feature matrix $\mathbf{Z} \in \mathbb{R}^{N \times F}$.

\subsubsection{Spatial Message Passing:}  
Spatial dependencies among channels are modeled using a single graph convolutional layer that implements message passing \cite{kipf2016semi}. The propagation rule is:
\begin{equation}
\mathbf{Z}^{(l+1)} = \sigma\Big( \tilde{\mathbf{D}}^{-1/2} \tilde{\mathbf{A}} \tilde{\mathbf{D}}^{-1/2} \mathbf{Z}^{(l)} \mathbf{W}^{(l)} \Big),
\label{eq:gcn}
\end{equation}
where $\tilde{\mathbf{A}} = \mathbf{A} + \mathbf{I}$ is the adjacency matrix with added self-loops, $\tilde{\mathbf{D}}$ is the diagonal degree matrix of $\tilde{\mathbf{A}}$, $\mathbf{Z}^{(l)}$ is the node feature matrix at layer $l$ (with $\mathbf{Z}^{(0)} = \mathbf{Z}$), $\mathbf{W}^{(l)}$ is the learnable weight matrix at layer $l$, and $\sigma(\cdot)$ is a ReLU activation function.

This process aggregates information from the neighbourhood of each node, capturing inter-electrode relationships.

After the spatial message passing, a global average pooling operation compresses the per-node features into a single trial-level feature vector $\mathbf{v} \in \mathbb{R}^{F'}$. Finally, the vector $\mathbf{v}$ is forwarded through a dense layer with ReLU activation and a sigmoid-activated output layer for binary classification.

\begin{figure}[h]
    \centering
    \includegraphics[width=0.4\linewidth]{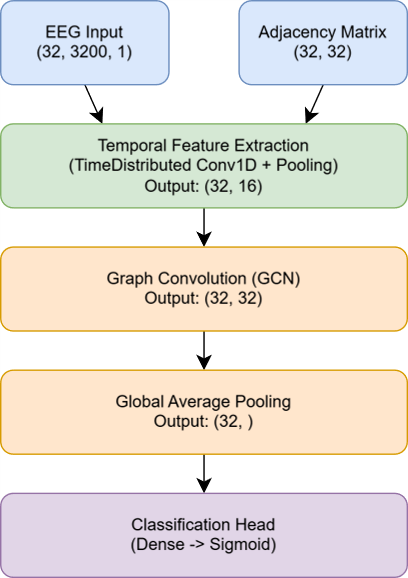}
    \caption{Schematic overview of the ST-GCN architecture. The network first extracts temporal features from each electrode via a TimeDistributed convolutional layer, followed by spatial message passing using a GCN layer that aggregates information via the propagation rule in Eq.~\ref{eq:gcn}. A global pooling layer then produces a trial-level feature vector for classification.}
    \label{fig:model_architecture}
\end{figure}

A schematic overview of the ST-GCN architecture is provided in Figure~\ref{fig:model_architecture}.

\section{Experimental Set-Up}
\label{sec:set-up}
\subsection{Dataset}
The SAM-40 dataset \cite{sam-40} consists of EEG recordings from 40 healthy subjects who completed a relaxation condition and three stress-inducing tasks. Each task is repeated over three seperate trials of 25 s each, recorded at 128 Hz, using 32 scalp electrodes (according to the International 10–20 system \cite{jasper1958ten_10_20_system}). In our experiments, we grouped the relaxation trials into a "Relaxed" class, while stress-inducing trials were grouped into a single "Stressed" class. This yields a total of 480 trials with 120 "Relaxed" elements and 360 "Stressed" elements. Each trial is represented by a 32 channels x 3200 time steps matrix. No additional filters were applied so that temporal features can be learned end-to-end. 

\subsection{Baseline Models}

The ST-GCN model is benchmarked against three baseline models:

\paragraph*{MLP:}  
A simple feedforward architecture designed to process flattened EEG input. Each input trial is reshaped into a vector of size $(32 \times 3200)$, passed through a dense layer with 64 hidden units and ReLU activation, followed by a dropout layer (rate = 0.3) to mitigate overfitting. A final dense layer with sigmoid activation produces the binary stress prediction. This architecture serves as a basic benchmark with no temporal or spatial inductive bias.

\paragraph*{LSTM:}  
A single-layer Long Short-Term Memory (LSTM) network to model temporal dependencies in the EEG signal. Input data is reshaped to $(3200, 32)$ per trial, treating each time step as a sequence with 32-dimensional input. The LSTM has 64 hidden units and is followed by a fully connected layer with a sigmoid output for binary classification. This setup captures sequential dynamics, making it more suited to EEG time-series than MLP.

\paragraph*{Transformer:}  
The Transformer baseline uses a lightweight encoder-only architecture to model long-range temporal dependencies. Each input trial has shape $(3200, 32)$. The architecture includes a single Transformer block with 4 attention heads and a feed-forward layer of size 64. Outputs are passed through residual connections, layer normalization, and dropout (rate = 0.1). Global average pooling condenses the sequence, followed by a dense hidden layer (32 units, ReLU) and a sigmoid output layer for classification. This model can effectively model temporal dependencies in the EEG signal.


\subsection{Implementation}
The code is implemented using TensorFlow and the spektral library. All baseline models (MLP, LSTM, Transformer) and the ST-GCN are evaluated under identical training conditions to ensure a fair comparison. All models use binary cross-entropy loss and the Adam optimiser, with a mini-batch size of 8 over 10 epochs, a validation split of 10\%, and a 0.001 learning rate. Data was split using stratified random sampling. Training was performed on Python 3.10.15 using an Intel i7-10510U CPU and an NVIDIA GeForce MX250 GPU. 

\subsection{Interpretability Analyses}

Understanding which aspects of the EEG signals drive model performance is important for scientific insight and model transparency. Therefore, we designed a set of ablation experiments that probe the contributions of spatial and temporal features in the ST-GCN's performance. Although these analyses do not provide definitive causal evidence for neurophysiological mechanisms, they offer valuable insights regarding the relative importance of individidual electrodes, brain regions, and time segments in the context of stress detection.

\subsubsection{Spatial Feature Importance:}  
We perform a single-channel ablation study where, for each trial, the signal from one electrode is preserved while signals from all other channels are set to zero. The resulting change in classification accuracy shows insights into the informativeness of each channel. In a complementary analysis, we group electrodes by brain region (e.g., frontal, central, parietal) and evaluate the impact on performance when: 
\begin{enumerate} 
    \item Only a specific region is used (i.e., all channels outside the region are zeroed). 
    \item A particular region is omitted while retaining all other channels. 
\end{enumerate} 

\subsubsection{Temporal Feature Importance:}  
To study the effect of individual time segments, the 3200 time steps are segmented into 10 equal parts (320 steps each). We perform two experiments: single-segment performance, where the model is trained using data from one time segment only, and segment removal, where one segment is removed from the input while retaining the remaining segments. The drop in accuracy reveals which time window is most critical for effective stress level detection.

\section{Results}
\label{sec:results}
The performance of the ST-GCN and benchmark models was evaluated on a binary classification task (Relaxed vs. Stressed) using raw EEG signals from the SAM-40 dataset \cite{sam-40}. Table~\ref{tab:results} summarises the accuracy, precision, recall, F1 score and AUC-ROC for the ST-GCN model compared with three baseline approaches: a Multilayer Perceptron (MLP), a Long Short-Term Memory (LSTM) network, and a Transformer-based model.

\begin{table}[]
    \centering
    \caption{Performance comparison of all models on the SAM-40 dataset. Accuracy, Precision, Recall, and F1 Score are reported as percentages; AUC-ROC is reported as a decimal. Values represent the mean and standard deviation over 10 runs.}
    \label{tab:results}
    \begin{tabular}{|l|c|c|c|c|c|}
    \hline
    \bfseries Model       & \bfseries Accuracy & \bfseries Precision & \bfseries Recall & \bfseries F1 Score & \bfseries AUC-ROC \\
    \hline
    MLP             & 55.63\% $\pm 4.40$ & 77.94\% $\pm 2.84$ & 57.08\% $\pm 6.98$ & 65.63\% $\pm 4.94$ & 0.56 $\pm 0.03$ \\
    LSTM            & 60.83\% $\pm 7.53$  & 76.20\% $\pm 1.77$ & 69.17\% $\pm 12.56$ & 71.98\% $\pm 7.78$ & 0.57 $\pm 0.05$ \\
    Transformer     & 61.25\% $\pm 11.26$  & 76.13\% $\pm 2.68$ & 69.72\% $\pm 20.72$ & 71.04\% $\pm 14.32$ & 0.54 $\pm 0.03$ \\
    ST-GCN & 69.06\% $\pm 8.27$  & 80.62\% $\pm 2.91$ & 78.33\% $\pm 17.48$ & 78.08\% $\pm 8.99$ & 0.67 $\pm 0.01$ \\
    \hline
    \end{tabular}
\end{table}

The ST-GCN model, which leverages both spatial and temporal dependencies in EEG, outperforms the MLP, LSTM, and Transformer baselines. Compared to the best baseline, ST-GCN’s mean accuracy is 8 percentage points higher, and it achieves an AUC-ROC of 0.67, indicating stronger overall performance. Crucially, this improvement is obtained using raw, unfiltered EEG, showing that integrating electrode geometry with temporal processing can extract informative stress‐related patterns without conventional band-pass filtering.

\subsection{Feature Importance Studies}

The graph-based approach to modelling the stressed brain additionally offers improved interpretability over standard deep learning models. By mapping nodes to electrodes in the standard 10–20 system, we can trace signal sources to specific electrode locations that correspond to distinct regions of the brain.

\subsubsection{Channel Importance}

Recent studies have identified specific EEG channels that are particularly sensitive to mental stress states. A channel selection method based on the correlation coefficient of Hjorth parameters identified eight general optimal channels (GOCs) for stress detection: AF3, FC5, F8, Fp1, AF4, P7, Fp2, and F7 \cite{hag2023mental_eeg_biomarkers}. These channels are predominantly located in the frontal lobe.

\begin{figure}[]
    \centering
    \includegraphics[width=0.5\linewidth]{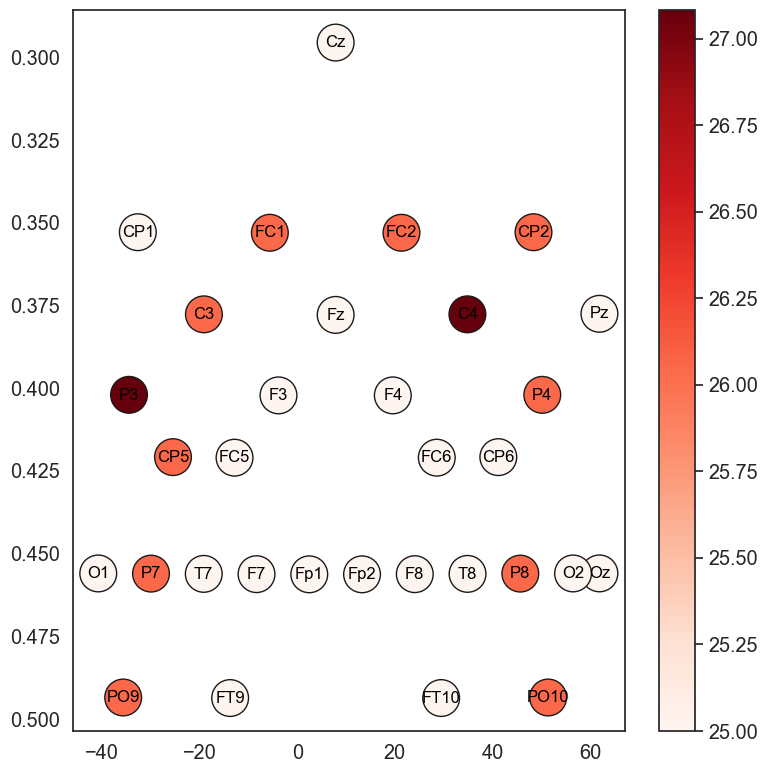}
    \caption{Topographical mapping of single-channel performance. Darker red indicates higher accuracy. Results suggests that P3 and C4 channels are among the largest contributors to effective stress detection.}
    \label{fig:channel-importance}
\end{figure}

When we feed the model only one electrode at a time, single-channel accuracy ranges from 25\% to 27\%, which are below the 50\% chance baseline, indicating no single electrode suffices for discrimination (See Figure \ref{fig:channel-importance}). Nonetheless, among these low scores, channels P3 and C4 achieve the highest accuracies, with other relatively higher scores seen at FC1, FC2, C3, CP2, and CP5. The results are consistent with the existing literature on stress biomarkers in EEG, emphasising the importance of frontal and central regions for stress processing \cite{alshorman2022frontal,xia2019physiological,yu2012estimating_flow_from_central_region_during_arithmetics}. Removing any single channel from the full model causes less than a 1\% drop in overall accuracy, indicating that stress-related information is distributed across multiple electrodes.

\subsubsection{Brain Region Ablation}
\begin{figure}[h]
    \centering
    \subfloat[Impact on accuracy by region removal.]{%
        \includegraphics[width=0.45\linewidth]{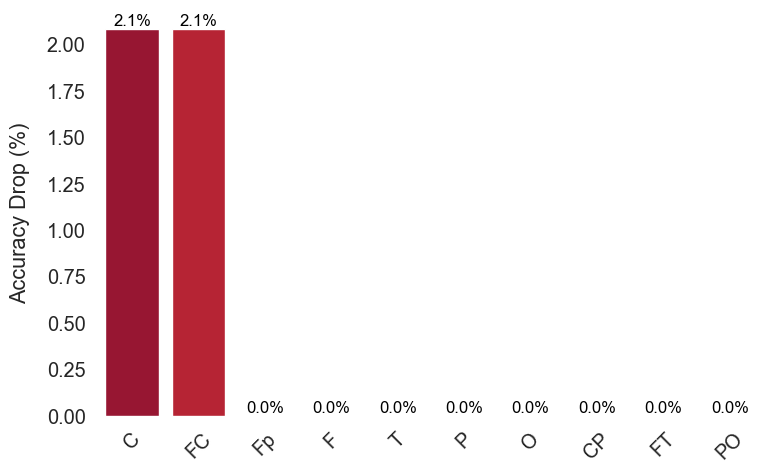}%
        \label{fig:region-ablation}%
    }
    \hfill
    \subfloat[Accuracy with individual brain regions only.]{%
        \includegraphics[width=0.45\linewidth]{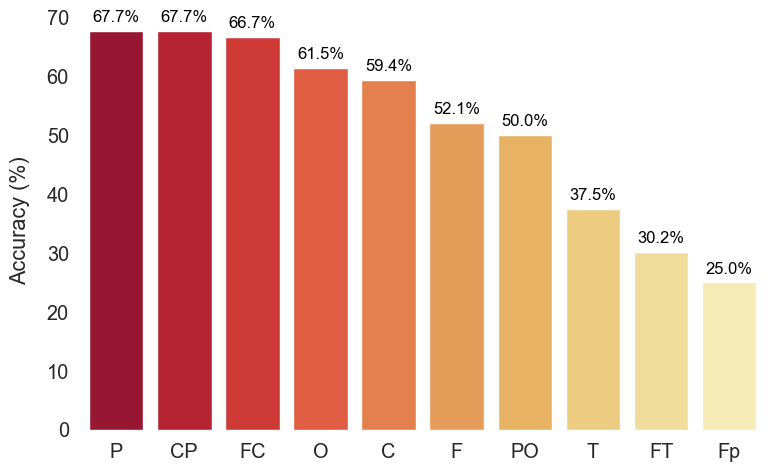}%
        \label{fig:accuracy-per-region}%
    }
    \caption{(a) Ablation study on brain regions. (b) Single-region performance.}
\end{figure}

Since removing any single channel caused less than a 1\% drop in accuracy, we next evaluated whole‐region ablation. Figure \ref{fig:region-ablation} (left) shows changes in accuracy when each predefined brain region is omitted. Excluding the central or frontal-central regions yields about a 2.1\% decrease in accuracy; removing other regions causes negligible drops (< 1\%). Conversely, when training on only one region at a time (Figure \ref{fig:accuracy-per-region}, right), the parietal and central-parietal regions achieve the highest accuracies ($\approx$ 68\%), well above the 50\% chance level, followed by the frontal-central region ($\approx$ 67\%). Frontal and frontal-temporal regions score around 25\% – 30\%, below chance. These results suggest that central and parietal electrodes hold the most stress-related information. The relatively weak performance of frontal cluster, which is associated with emotion and stress processing  \cite{alshorman2022frontal,xia2019physiological}, may reflect the noisy nature of raw EEG or could be due to the specifics of our graph construction. Future work should explore these possibilities further.

\subsubsection{Temporal Ablation Studies}

\begin{figure}[ht]
    \centering
    \subfloat[Accuracy when training with only one time segment.]{%
        \includegraphics[width=0.45\linewidth]{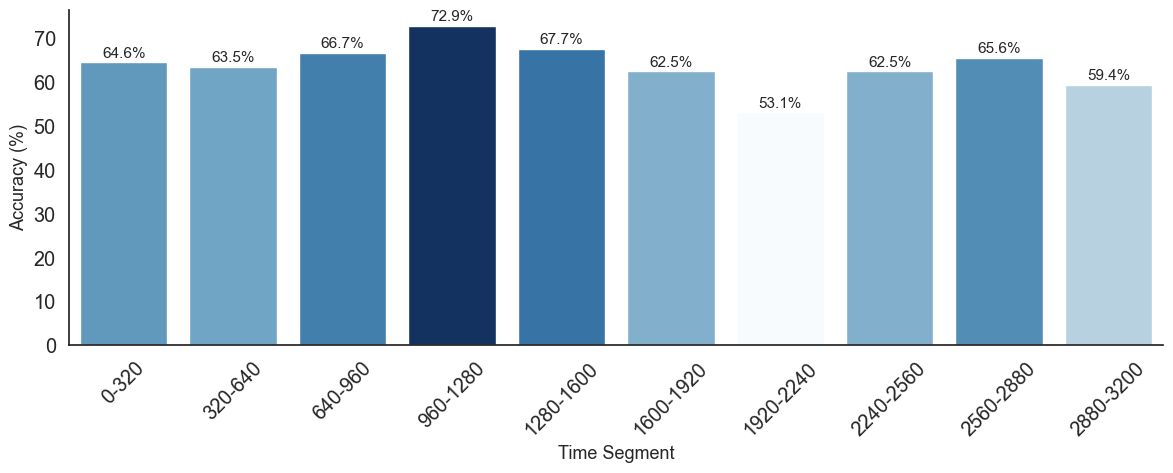}%
        \label{fig:accuracy-per-time-segment}%
    }
    \hfill
    \subfloat[Decrease in accuracy upon removal of a time segment.]{%
        \includegraphics[width=0.45\linewidth]{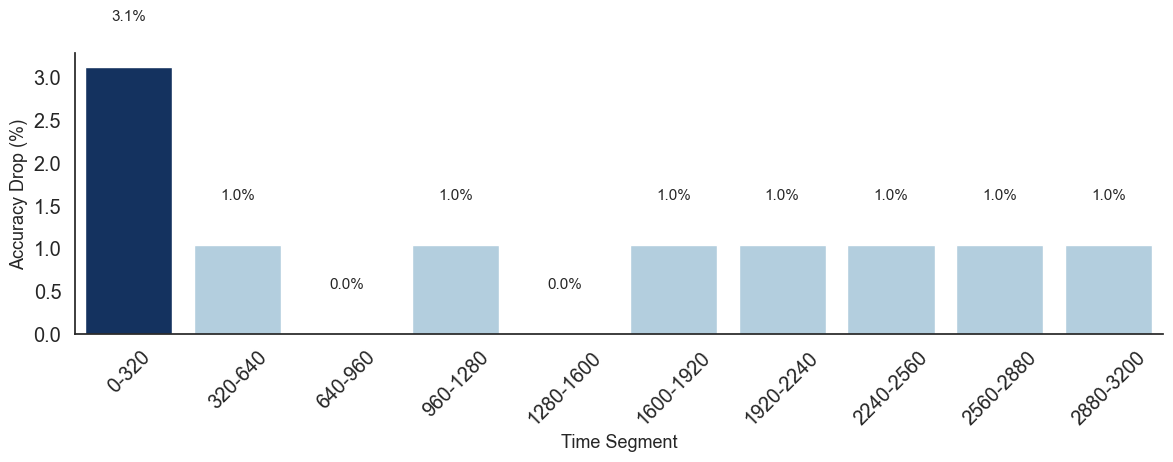}%
        \label{fig:time-ablation}%
    }
    \caption{Temporal ablation studies: (a) Isolated time segment performance; (b) Accuracy drop when a specific segment is removed.}
\end{figure}

To investigate how stress-related neural activity changes over time, we performed a set of temporal ablation experiments. Each trial, consisting of 3200 time steps, was partitioned into 10 non-overlapping segments of 320 steps. In the first experiment, we trained and evaluated the model using only a single time segment at a time. As shown in Figure \ref{fig:accuracy-per-time-segment}, the fourth segment yielded the highest stand-alone performance (73\% accuracy), while the seventh and tenth segments resulted in the weakest classification performance (53.1\% and 59.4\%, respectively).

In a complementary setup, we measured the decrease in overall model accuracy when each time segment was removed from the full input sequence (Figure \ref{fig:time-ablation}). The first segment proved to be the most critical, with its removal reducing accuracy by approximately 3 percentage points. In contrast, omitting segments three and five had negligible impact.

These results suggest that the early and mid-trial periods are particularly informative for stress classification, potentially capturing initial cognitive or physiological responses to task-induced stress. The reduced informativeness of later segments may reflect neural adaptation or reduced signal-to-noise ratio as the trial progresses.

\section{Conclusion and Future Work}
\label{sec:conclusion}
In this work, we introduced a novel EEG‐based stress detection framework that converts raw EEG signals into graphs and processes them with a spatio‐temporal graph‐convolutional network (ST‐GCN). By leveraging both anatomical and functional connectivity, our model outperforms traditional baselines (MLP, LSTM, Transformer) on accuracy, F1 score, and AUC‐ROC.

However, there are important limitations. Our evaluation was restricted to the SAM‐40 dataset (40 participants), which may limit generalisability. Using unfiltered EEG, while intentionally stressing end‐to‐end learning, may have added noise and impaired feature extraction. Moreover, our ablation studies revealed an unexpectedly low contribution from frontal lobe electrodes, diverging from prior literature and possibly reflecting noise or other factors; this discrepancy should be investigated further.

Future research should validate this framework on larger, more diverse datasets and explore standard preprocessing (e.g., band-pass filtering, artefact removal) to reduce noise. Investigating more advanced graph‐based architectures (inspired by recent work on EEG emotion recognition) may better capture complex EEG patterns. Finally, deeper interpretability studies are needed to explain the frontal lobe anomaly and identify reliable neural biomarkers of stress.

In summary, our work demonstrates the potential of graph‐based approaches for EEG‐based stress detection. Addressing these limitations, especially by expanding datasets, reducing noise, and performing more in‐depth interpretability analyses, will be crucial for developing robust, interpretable stress monitoring systems.


%
%
%
\bibliographystyle{splncs04}

%




\end{document}